\documentclass[letterpaper, 10 pt, conference]{ieeeconf}  

\IEEEoverridecommandlockouts                              

\usepackage{cite}
\usepackage[hyphens]{url}
\usepackage{hyperref}
\usepackage{romannum}
\usepackage{amsmath,amssymb,amsfonts}
\usepackage{algorithmic}
\usepackage[ruled,vlined]{algorithm2e}
\usepackage{graphicx}
\usepackage{booktabs}
\usepackage{textcomp}
\usepackage{xcolor}
\usepackage{subfig}
\usepackage{bm}
\usepackage{siunitx}
\usepackage{tabularx} 

\title{\LARGE \bf
Learning Robust Control Policies for Inverted Pose on Miniature Blimp Robots
}

\author{Yuanlin Yang, Lin Hong, and Fumin Zhang$^\dagger$
 \thanks{The work described in this paper was partially supported by grants AoE/E-601/24-N, 16203223, N\_HKUST677/24, C6029-23G, and C6078-25G from the Research Grants Council of
 the Hong Kong SAR, China.}
 \thanks{Yuanlin Yang, Lin Hong, and Fumin Zhang are with the Department of Electronic and Computer Engineering, The Hong Kong University of Science and Technology, Hong Kong, China. E-mails: 
         {\tt\small yyanghy@connect.ust.hk, eelinhong@ust.hk, eefumin@ust.hk}}%
         \thanks{$^\dagger$Corresponding author: Fumin Zhang (email: {\tt\small eefumin@ust.hk})}
}

\begin{document}

\maketitle
\thispagestyle{empty}
\pagestyle{empty}

\begin{abstract}
The ability to achieve and maintain inverted poses is essential for unlocking the full agility of miniature blimp robots (MBRs). However, developing reliable inverted control strategies for MBRs remains challenging due to their complex and underactuated dynamics. To address this challenge, we propose a novel framework that enables robust control policy learning for inverted pose on MBRs. The proposed framework consists of three core stages. First, a high-fidelity three-dimensional (3D) simulation environment is constructed and calibrated using real-world MBR motion data. Second, a robust inverted control policy is trained in simulation using a modified Twin Delayed Deep Deterministic Policy Gradient (TD3) algorithm combined with a domain randomization strategy. Third, a mapping layer is designed to bridge the sim-to-real gap and facilitate real-world deployment of the learned policy. Comprehensive evaluations in the simulation environment demonstrate that the learned policy achieves a higher success rate compared to the energy-shaping controller. Furthermore, experimental results confirm that the learned policy with a mapping layer enables an MBR to achieve and maintain a fully inverted pose in real-world settings.
\end{abstract}

\section{INTRODUCTION}
The potential of agile flying in Unmanned Aerial Vehicles (UAVs) has been extensively demonstrated using proportional–integral–derivative (PID) control~\cite{wang2025unlocking}, model predictive control (MPC)~\cite{torrente2021data}, and deep reinforcement learning (DRL)~\cite{xie2023learning,han2025reactive}. However, for MBRs, a distinct category of aerial platforms, a significant gap remains in the development of advanced control strategies capable of delivering comparable agility. UAVs typically rely on high-speed rotating propellers for lift and maneuvering, which inherently results in high energy consumption and potential safety risks when operating in proximity to humans. In contrast, MBRs utilize buoyant gas to offset their weight and employ low-power thrusters for fine-grained motion control.
This unique design has positioned MBRs as a promising solution for various applications, such as entertainment and advertising \cite{oh2006flying}, warehouse inventory management~\cite{han4988309flying}, indoor environmental monitoring~\cite{chaitanya2025slam}, and infrastructure inspection~\cite{nitta2017visual}.

Existing research on MBRs has predominantly centered on innovative structural design, including optimization of the envelope shape, gondola layout, and payload integration to enhance operational stability and adaptability~\cite{zhang2023novel,ye2022design}. Control studies have mainly addressed small pitch or yaw adjustments for basic hovering~\cite{xu2023sblimp} or low-speed navigation~\cite{tao2021swing}. However, fully agile motion control of MBRs, a capability that would enable rapid attitude transitions and wide-range position adjustments, remains an open challenge.
This challenge arises from the unique dynamic properties of MBRs. In low-speed small UAV applications, aerodynamic drag is typically negligible relative to total thrust, as UAVs rely on high-power propellers to counteract their full weight; consequently, their thrust output significantly exceeds drag forces during low-speed motion. For MBRs, two key characteristics reverse this relationship:
(1) Dominant aerodynamic drag due to their large envelope volume;
(2) Weak thrust output, since buoyant gas offsets most of their weight, eliminating the need for high thrust to counteract gravity.

These distinct dynamic properties make MBR attitude control fundamentally different from that of UAVs, rendering conventional UAV control strategies largely inapplicable. In addition, a review of MBR designs~\cite{cho2017autopilot, cheng2023rgblimp, xu2025mochiswarm, pellegrino2024tinyblimp, zhang2023novel, ye2022design} further highlights a structural constraint: most MBRs adopt a gondola-envelope configuration, where the gondola, housing sensors, thrusters, and controllers, is suspended below or attached to the envelope. This structure inherently exhibits both stable and unstable equilibrium points: for example, the ``upright" pose (gondola hanging below the envelope) is a stable equilibrium, while the ``inverted" pose (gondola above the envelope) is unstable and difficult to maintain.
\begin{figure}[t]
    \centering
    \includegraphics[width=1\linewidth]{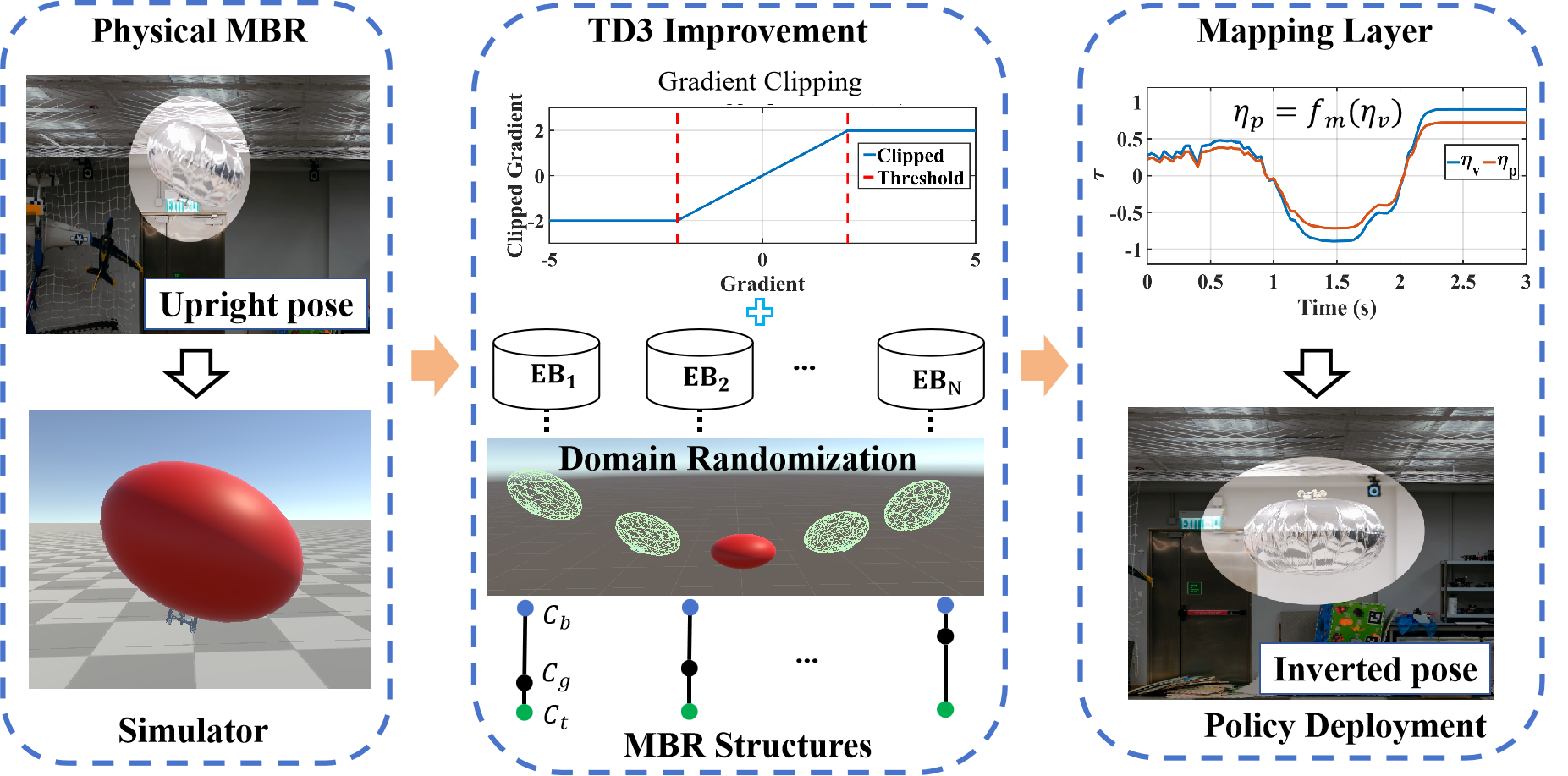}
    \caption{Overview of the proposed method for inverted pose on MBRs. The MBR with an upright pose can reach and maintain the inverted pose with the learned robust policy.}
    \label{fig:general_method_framework}
\end{figure}

Against this backdrop, the primary task in this paper is to enable MBRs to reach and maintain an inverted pose based on DRL, as illustrated in Fig.~\ref{fig:general_method_framework}. 
To clarify the problem formulation, we define \emph{inverted control} as the ability of an MBR to achieve and stabilize a fully upside-down pose—corresponding to an unstable equilibrium state in which the center of buoyancy lies below the center of gravity.
The work most closely related is a recent study by Wang and Zhang~\cite{wang2024achieving}, which explicitly tackles the challenge of inverted control for MBRs. Their approach successfully demonstrates the achievement and maintenance of a stable inverted pose using model-based control: an energy-shaping controller to tailor the system’s energy landscape for state transitions, paired with a linear state feedback controller to suppress deviations from the target inverted state. However, the energy calculation at the core of their controller depends on time-invariant MBR dynamics, yet the model parameters are highly dynamic in real-world operations, leading to performance degradation or even loss of inverted stability under environmental disturbances.
Recent progress in DRL-based control has shown promise for addressing the parameter variability and disturbance susceptibility of outdoor large-size blimp robots. Liu et al. \cite{liu2022deep} proposed a deep residual reinforcement learning method integrated with a PID controller in a closed loop. This hybrid approach improved trajectory tracking accuracy but remained limited to small-range attitude control, with no consideration of inverted states. Zou et al. \cite{zuo2023autonomous} designed a hybrid control framework that combines robust $H_\infty$ control with proximal policy optimization (PPO), which enhances robustness against wind disturbances and variations in buoyancy. 

Despite these advances, developing DRL-based methods for inverted control of MBR remains largely unexplored. 
In this paper, we address this gap by presenting a robust policy specifically designed for inverted control of MBRs. Our approach combines domain randomization, multi-buffer experience replay, and a sim-to-real transfer strategy to expand the applicability of learning-based control to MBRs. By focusing on inverted control as a cornerstone of large-envelope agility, this study aims to unlock new capabilities for MBRs.
The main contributions are as follows:
\begin{itemize}
\item To the best of our knowledge, this work presents the first Unity-based 3D simulation environment~\cite{Unity2025} specifically designed for inverted control of MBRs. The simulator captures MBR-specific dynamics and enables diverse scenario generation for robust policy training.

\item We propose a learning framework for robust inverted control of MBRs. The framework integrates domain randomization to improve robustness against parameter variations and disturbances, and introduces refinements to TD3 to enhance training stability.

\item We develop a sim-to-real transfer strategy with a mapping layer to compensate for discrepancies between simulated and physical dynamics. Experimental results demonstrate that the learned policy reliably achieves inverted stabilization on a real MBR without additional policy retraining.
\end{itemize}

\section{Problem Formulation}
\subsection{Dynamic Model of MBRs}\label{st:blimp_system}
\begin{figure}[t]
    \centering
    \includegraphics[width=1\linewidth]{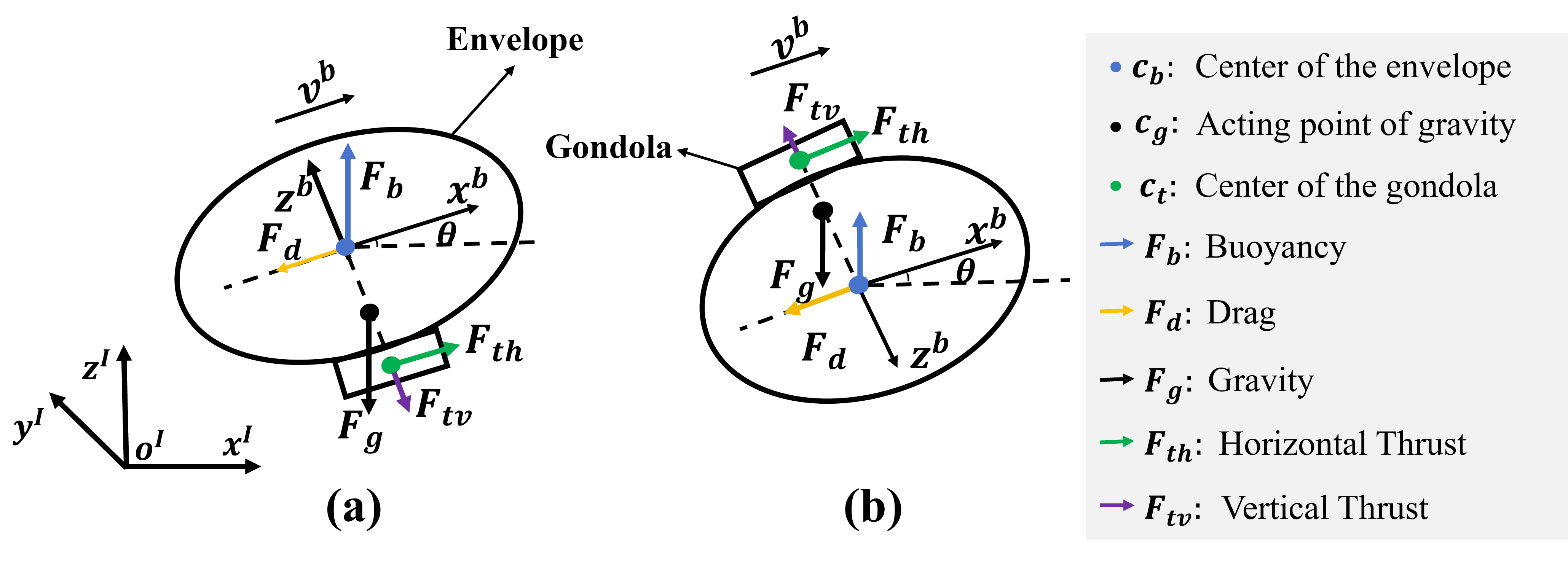}
    \caption{Dynamic model analysis of the MBR: (a) Upright pose; (b) Inverted pose.}
    \label{fig:structure_force}
\end{figure}

As shown in Fig.~\ref{fig:structure_force}, the MBR consists of an envelope and a gondola: the envelope is to supply the buoyancy, while the gondola is to provide a platform for housing thrusters and other electrical devices.
Based on the first principle and~\cite{tao2021swing, dong2024adaptive}, the dynamic model of the MBR can be expressed as:
\begin{equation} \label{eq:dynamic_model}
(\bm{M_{rb}+\bm{M_{a}}})\dot{\bm{\nu}}_{b/e}^b + (\bm{C_{rb}}+\bm{C_{a}})\bm{\nu}_{b/e}^b + \bm{\Gamma}^d + \bm{\Gamma}^{gb} + \bm{\Gamma}^{e}= \bm{\Gamma}^t,
\end{equation}
where $\bm{\Gamma}^d$ denotes air drag, 
$\bm{\Gamma}^t$ and $\bm{\Gamma}^{e}$ correspond to thruster-generated and environmental forces and torques, respectively. $\bm{\Gamma}^{gb}$ involves the restoring force and torque.

The dynamic models near the upright and inverted poses are shown in Figures~\ref{fig:structure_force} (a) and (b), respectively. In both cases, the MBR operates at a constant velocity while holding a stable attitude. Based on the force analysis, we have
\begin{equation} \label{eq:inv_up_dynamic_model}
    \mathbf{F}_{th}r_t^b + m_{rb}\mathbf{g}r_z^b\sin(\theta) = \mathbf{0},
    \mathbf{F}_b + \mathbf{F}_g = \mathbf{0},
    \mathbf{F}_{th} + \mathbf{F}_d = \mathbf{0},
\end{equation}
where $r_t^b$ is the distance between $c_b$ and $c_t$, $r_z^b$ is the distance between $c_b$ and $c_g$, and $m_{rb}$ is the MBR total mass. 
\begin{figure*}[t]
    \centering
    \includegraphics[width=1\linewidth]{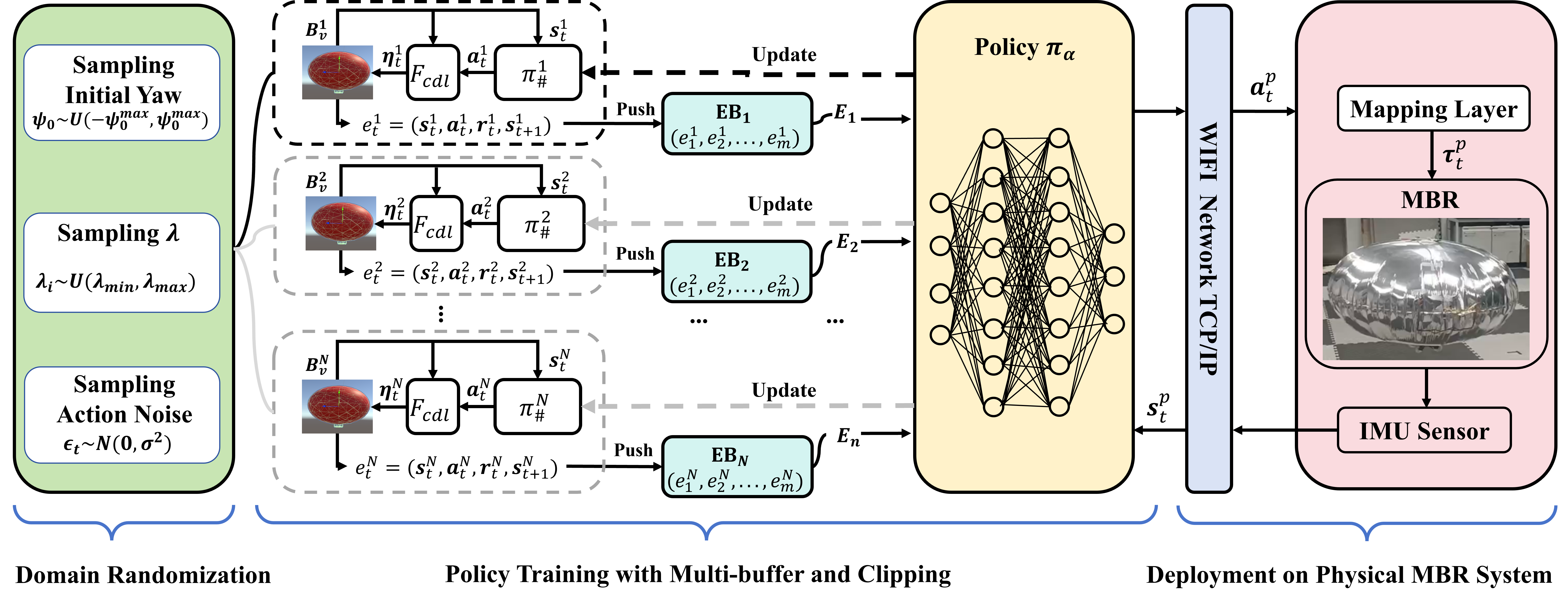}
    \caption{Pipeline of developing and deploying robust policy for inverted control of MBRs.}
    \label{fig:system_framework}
\end{figure*}
\subsection{Inverted Control Problem Statement}
According to (\ref{eq:dynamic_model}) and (\ref{eq:inv_up_dynamic_model}), the MBR exhibits highly nonlinear dynamics, and its attitude control performance is sensitive to variations in model parameters. The objective of this study is to design a robust control policy, denoted as $\pi_{\alpha}$, that drives the MBR from its stable equilibrium state ($\bm{\Theta}_0$) to the unstable equilibrium state ($\bm{\Theta}_d$), and maintains stabilization in its vicinity.
The orientation of the MBR is represented as $\bm{\Theta} = [\phi, \theta, \psi]^T$, where $\phi$, $\theta$, and $\psi$ denote roll, pitch, and yaw angles, respectively. The control objective can therefore be formulated as:
\begin{equation}
    \lim_{t\rightarrow \infty}||\mathbf{\Theta}_d-\mathbf{\Theta}_t||_{M_{{\Theta}}} = 0, \mathbf{\Theta}_t = T_b(\mathbf{\Theta}_{t-1}, \boldsymbol{a}_t, \mathbf{S}_b),
\end{equation}
where $T_b$ is the dynamics transition from $\mathbf{\Theta}_{t-1}$ to $\mathbf{\Theta}_t$ under the control command $\boldsymbol{a}_{t}$ generated by the designed policy $\pi_{\alpha}$. $\mathbf{S}_b$ involves the parameters in the MBRs' dynamic model.

The policy $\pi_{\alpha}$ aims to maximize the total cumulative reward the MBR receives over the long run through interaction with the environment, formulated as:
\begin{align}
     & \max_{\alpha} \quad G_t \\
    \text{s.t.} \quad &G_t = \sum_{k=0}^{\infty}\gamma^kr_{t+k+1}, r_{t} = f_r(\mathbf{s}_t,\mathbf{s}_{t+1}, \boldsymbol{a}_{t}),\\
                      & s_{t+1} = T_b^{v}(\mathbf{s}_t, \boldsymbol{a}_{t}, \mathbf{S}_b^v), \boldsymbol{a}_{t} = \pi_{\alpha}(\mathbf{s}_t),
\end{align}
where $G_t$ is the cumulative reward, $\gamma$ is the discount factor, and $f_r$ is the designed reward function to evaluate the action $\boldsymbol{a}_{t}$ taken in state $\mathbf{s}_t$. $T_b^v$ and $\mathbf{S}_b^v$ represent the dynamics transition and the parameters of the MBR in the simulation environment, respectively.

Due to unmodeled dynamics and parameter mismatches in the training environment, bridging the gap between the simulated environment and the physical setting needs to be considered in the policy deployment phase. Assuming there is a mapping function $f_m$ that can bridge the gap, the problem is formulated as:
\begin{align}
    & \lim_{t\rightarrow \infty}||\mathbf{s}_d-\mathbf{s}_{t}||_{M_s} = 0, \\
    \mathbf{s}_{t+1} &= T_b^p(\mathbf{s}_{t-1}, \boldsymbol{a}^p_{t}, \mathbf{S}_b^p), \boldsymbol{a}^p_{t} = f_m(\pi_{\alpha}(\mathbf{s}_t)),
\end{align}
where the dynamics transition and parameters in the physical system are different from those in the simulated environment, $T_b^p \neq T_b^v$ and $\mathbf{S}_b^v \neq \mathbf{S}_b^p$. $\mathbf{s}_d$ denotes the desired state. 

\section{Methodology}
The pipeline of learning a robust policy for inverted control of MBRs is illustrated in Fig.~\ref{fig:system_framework}, which incorporates three core stages: (1) 3D simulation environment creation, (2) Physics-informed domain randomization strategy design, and (3) TD3 with multi-buffer and clipping.


\subsection{Simulation Environment}
\begin{figure}[t]
    \centering
    \includegraphics[width=1\linewidth]{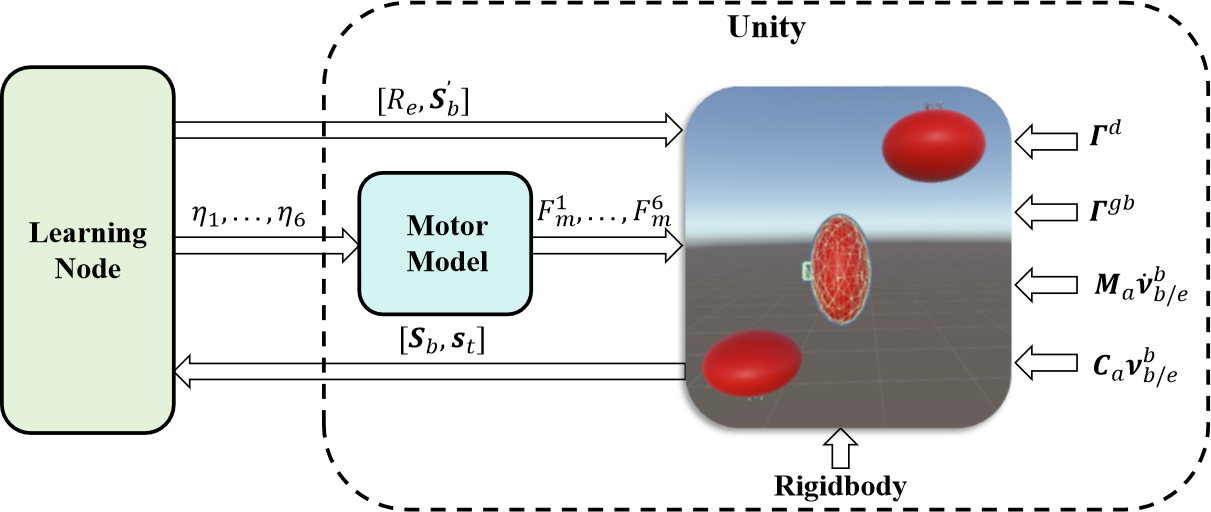}
    \caption{3D simulation environment designed for robust policy learning for inverted control of MBRs.}
    \label{fig:learning_env}
\end{figure}
As illustrated in Fig.~\ref{fig:learning_env}, 
Unity was adopted as the simulation platform to implement the MBR dynamics and construct the policy training environment. The Rigidbody component was used to reproduce the dynamics. Custom force and torque terms were implemented via the APIs AddForceAtPosition and AddRelativeTorque, including aerodynamic drag $\bm{\Gamma}^{d}$, restoring force and torque $\bm{\Gamma}^{gb}$, and added-mass and added-inertia effects ($\mathbf{M}_a$, $\mathbf{C}_a$).
Model parameters were identified following \cite{tao2018parameter,tao2020modeling}. To improve suitability for inverted control training, three enhancements were introduced.
First, a refined motor model was developed using experimental data and calibrated according to \cite{tao2020design}:
\begin{equation}
F_m = g_m(-0.0292\eta^2 + 0.1118\eta - 0.0039),
\end{equation}
where $g_m$ is the motor gain and $\eta \in [0,1]$ the control input. Varying $g_m$ enables simulation of actuator variability.

\begin{figure}[t] \centering \includegraphics[width=1\linewidth]{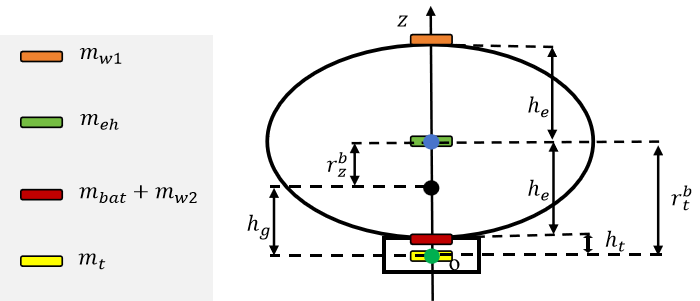} \caption{The MBR's structure in the simulation environment.} \label{fig:blimp_structure_2} \end{figure}

Second, the simulated MBR structure was modified (Fig.~\ref{fig:blimp_structure_2}) by decomposing the total additional mass into two components, $m_{w_1}$ and $m_{w_2}$, facilitating inverted control training.
Third, a Python-based learning node was implemented to manage training interaction. It supports episode reset via the command $R_e$ and online configuration of MBR parameters $\mathbf{S}_b^{'}$. The variables $\mathbf{S}_b$ and $\mathbf{s}_t$ denote the real-time system parameters and state, respectively.

\subsection{Physics-informed Domain Randomization}

As analyzed in Section \ref{st:blimp_system}, the distances between the three center points ($c_b$, $c_g$, and $c_t$) play a dominant role in the dynamics of the MBR. Accordingly, the proposed domain randomization strategy perturbs these distances while preserving physical consistency. As shown in Fig.~\ref{fig:blimp_structure_2}, the resultant gravitational force acts at $c_g$. The distance between $c_g$ and $c_t$ is
\begin{equation}
    h_g = \frac{(m_{bat} + m_{w2})h_t + (m_e + m_h)r_t^b + m_{w1}(2h_e + h_t)}{m_{rb}},
\end{equation}
where the total mass $m_{rb} = m_t + m_{bat} + m_e + m_h + m_{w1} + m_{w2}$. $m_t$ and $m_{bat}$ are the weights of the gondola and the battery, respectively. $m_w = m_{w1} + m_{w2}$ is the extra weight to ensure that the MBR is in a neutrally buoyant state. $m_e$ is the weight of the envelope in the deflated state, while $m_h$ is the weight of the filled helium, calculated by $m_h = \rho_{h}V$, where $\rho_h$ is the density of helium and $V$ is the total volume of the inflated envelope. The distance between $c_g$ and $c_b$ is $r_z^b = r_t^b - h_g$, where $r_t^b = h_t + h_e$. $h_t$ and $h_e$ are half-heights of the gondola and the inflated envelope, respectively. Denote $m_{w1} = \lambda m_w$ and $m_{w2} = (1-\lambda) m_w$; $h_g$ can be expressed in a more simplified way as 
\begin{equation}
    h_g = \frac{d_m + m_w(h_t + 2\lambda h_e)}{m},
\end{equation}
where $d_m = m_{bat}h_t + (m_e + m_h)r_z^b$. Adjusting $m_w$ and $\lambda$ can modify the distances between these center points. The key distinction is that only varying $\lambda$ allows $m_w$ to remain constant while altering $h_g$. 
\subsection{TD3 with Multi-buffer and Clipping}
TD3~\cite{fujimoto2018addressing} consists of two interrelated processes: environment interaction and policy optimization. During environment interaction, trajectories of the MBR are sampled under different actions, and the outcome of each action is evaluated through a corresponding reward signal.
As described in Algorithm~\ref{al:rl_experience_collection}, $N$ replay buffers are constructed to store MBR trajectories generated under different values of $\lambda$. 
\begin{algorithm}[t] 
\caption{Environmental Interaction with Domain Randomization}
\label{al:rl_experience_collection}
\KwData{Behavior policies $\pi_{\#}$, $N$ replay buffers $\mathcal{B}_1, \dots, \mathcal{B}_N$, running time of each episode $t_e$, maximum episode $N_e$, gaussian noise parameters $\sigma$, $\xi$ and $n_{\sigma}$, $N$ variable parameters $\lambda_1, \dots, \lambda_N$, $k=0$}
\KwResult{$N$ replay buffers with experiences}
\BlankLine

\For{$i = 1$ to $N_e$}{
    \If{$i \bmod n_{\sigma} = 0$}{
        $\sigma \leftarrow \xi \sigma$ 
    }
    $t_s \leftarrow$ \text{GetCurrentSystemTime}
    
    $\psi^i \leftarrow U(-\psi_0, \psi_0)$
    
     \text{UpdateBlimpParameter}($\lambda_k$, $\psi^i$)
     
    \While{True}{
        $\epsilon_t^k \sim \mathcal{N}(0, \sigma^2)$
        
        Observe state $s_t^k$, select action $a_t^k = \pi_{\#}^t(s_t^k) + \epsilon_t^k$
        
        $\phi_t^k, \theta_t^k \leftarrow \text{GetAttitudeAngles($s_t^k$)}$
        
        $\mathbf{\eta_t^k} \leftarrow F_{cdl}(a_t^k, \phi_t^k, \theta_t^k)$
        
        Execute $\mathbf{\eta_t^k}$, observe reward $r_t^k$, next state $s_{t+1}^k$, done flag $d_t^k$
        
        Store $(s_t^k, a_t^k, r_t^k, s_{t+1}^k, d_t^k)$ in $\mathcal{B}_k$
        
        $k \leftarrow k+1$
        
        \If{$k == N$}{
        $k = 0$
        
        }
        $t_c \leftarrow$ \text{GetCurrentSystemTime}
        
        \If{$t_c - t_s \leq t_e$ or \text{over range}}{
            \text{break}
        }
    }
}
\end{algorithm}


Once the replay buffers are sufficiently populated, a separate training thread is initiated to update the policy, as detailed in Algorithm~\ref{al:rl_policy_training}.
\begin{algorithm}[t]
\caption{TD3 with Multi-Buffer and Clipping}
\label{al:rl_policy_training}
\KwData{$N$ replay buffers $\mathcal{B}_1, \dots, \mathcal{B}_N$, interval of policy delay update $d_p$, discount $\gamma$, target network update rate $\varpi$, thresholds for gradient clipping $c_\alpha$ and $c_\beta$, actor $\pi_{\alpha}$, critics $Q_{\beta_1}, Q_{\beta_2}$, target networks $\pi_{\alpha'}, Q_{\beta'_1}, Q_{\beta'_2}$}
\KwResult{Learned policy $\pi_{\alpha}$}
\BlankLine
\While{True}{
    \For{buffer $k = 1$ \KwTo $N$}{
        Sample batch $(s_i^k, a_i^k, r_i^k, s_{i+1}^k, d_i^k) \sim \mathcal{B}_k$
    }
    \textbf{Critics Update:}
    Aggregate target Q-values: $Q_{\text{d}} = \frac{1}{N} \sum\limits_{k=1}^N \left[r_i^k + \gamma (1 - d_i^k) \min\limits_{j=1,2} Q_{\beta'_j}(s_{i+1}^k, \pi_{\alpha'}(s_{i+1}^k))\right]$
    
    Update gradients ($j=1,2$): $\nabla_{\beta_j} = \frac{1}{N} \sum_{k=1}^N \sum_i (Q_{\beta_j}(s_i^k, a_i^k) - Q_{\text{d}})^2$
    
    Clip gradients: $\text{clip}(\nabla_{\beta_j}, -c_\beta, c_\beta)$
    
    \textbf{Delayed Actor Update:}
    
    \If{$t \bmod d_p = 0$}{
        Update gradients $\pi_{\alpha}$: $\nabla_{\alpha} = \frac{1}{N} \sum_{k=1}^N \sum_i Q_{\beta_1}(s_i^k, \pi_{\alpha}(s_i^k))$
        
        Clip gradients: \text{clip}$(\nabla_{\alpha}, -c_\alpha, c_\alpha)$
        
    }
    \textbf{Target Network Updates:}
    
    Soft update: $\beta'_j \leftarrow \varpi \beta_j + (1 - \varpi) \beta'_j$ for $j=1,2$
    
    Soft update: $\alpha' \leftarrow \varpi \alpha + (1 -\varpi) \alpha'$
}
\end{algorithm}
The overall training framework follows the standard TD3 architecture and incorporates gradient clipping operations ($c_{\alpha}$ and $c_{\beta}$), adopted from PPO~\cite{schulman2017proximal}, to further improve training stability. Instead of updating the policy using a single replay buffer, the proposed method leverages $N$ distinct replay buffers, each containing trajectories generated under different MBR dynamic configurations. This multi-buffer training strategy encourages the policy to learn more generalized features, thereby improving its robustness across a wide range of dynamic conditions.

The state representation of the MBR consists of the rotation matrix $\mathbf{R}$ and the angular velocity vector $\boldsymbol{\omega}$. The action space is defined as the desired control torques about the three rotational axes. These torques are subsequently mapped to motor commands through the functional module $F_{cdl}$ shown in Fig.~\ref{fig:system_framework}, implemented following the method described in~\cite{tao2020design}.










The reward function comprises three components: an orientation reward $r_{rot}$, an angular velocity cost $r_{\omega}$, and an action cost $r_{a}$, which can be expressed as
\begin{equation}\label{eq:reward_function}
    r_{total} = r_{rot} + r_{\omega} + r_{a},
\end{equation}
where $r_{\omega} = -\frac{\sum_{i \in \{x,y,z\}}g_{\omega_i} |\omega_i|}{\omega_{max}}$. The parameter $g_{\omega_i}$ represents the weight assigned to each channel, and $\omega_{max}$ denotes the maximal angular velocity. The action cost is denoted by $r_a = -\sum_{i \in \{x,y,z\}} g_{ai}|\tau_i|$, where $g_{ai}$ is the parameter to shape the reward, and $\tau_i$ represents the expected torque of the $i$-th axis. The orientation reward is defined as
\begin{align}
r_{rot} &= \exp(-\text{clip}(g_{\phi}|e_{\phi}| + g_{\theta}|e_{\theta}| + g_{\psi}|e_{\psi}|, 0, g_n)) \\  \nonumber
        &+ \mathbb{I}_{\{\varphi < \zeta\}} \cdot \left(1 - \frac{\varphi}{\zeta}\right),
\end{align}
where $e_{\phi} = \frac{\varphi \mathbf{v}_x}{\pi}$, $e_{\theta} = \frac{\varphi \mathbf{v}_y}{\pi}$ and $e_{\psi} = \frac{\varphi \mathbf{v}_z}{\pi}$. The pair $(\mathbf{v}, \varphi)$ represents the axis-angle parameterization of the orientation error $\mathbf{R_e} = \mathbf{R}^{T}\mathbf{R}_d$, where $\mathbf{R}_d = \text{diag}(1, -1, -1)$. The rotation angle $\varphi$ is calculated by $\varphi = \arccos{(\min(\max(\frac{tr(\mathbf{R}_e) - 1}{2}, -1), 1))}$, and the rotation axis is given by: $\mathbf{v} = \frac{1}{2\sin\varphi}\begin{bmatrix}
    \mathbf{R}_{e32} - \mathbf{R}_{e23} \\
    \mathbf{R}_{e13} - \mathbf{R}_{e31} \\
    \mathbf{R}_{e21} - \mathbf{R}_{e12}
\end{bmatrix}$. 
The symbol $\mathbb{I}$ denotes the indicator function where the value is equal to 1 if the condition is true, 0 otherwise. The parameter $\zeta$ represents the threshold for the precision bonus.

\subsection{Implementation Details}
Both the policy $\pi_{\alpha}$ and the action-value function $Q_{\beta}$ were approximated using fully connected neural networks with two hidden layers of 256 neurons each. All hidden layers employ Leaky ReLU activation functions, while the output layer of the policy uses a hyperbolic tangent (Tanh) activation.
The reward function parameters are summarized in Table~\ref{tab:reward_params}. Higher weights are assigned to $\phi$ and $\theta$ deviations than to $\psi$, reflecting the priority of inverted pose stabilization.
\begin{table}[t]
    \centering
    \setlength{\tabcolsep}{4.5pt}  
    \caption{Parameters in the reward function}
    \label{tab:reward_params}
    \begin{tabular}{*{11}c}
        \toprule
        $g_{\omega_x}$ & $g_{\omega_y}$ & $g_{\omega_z}$ & 
        $g_{a_x}$ & $g_{a_y}$ & $g_{a_z}$ & 
        $g_\phi$ & $g_\theta$ & $g_\psi$ & $\zeta$ & $g_{n}$\\
        \midrule
        0.01 & 0.01 & 0.01 & 
        0.001 & 0.001 & 0.001 & 
        5.0 & 5.0 & 0.5 & 0.1 & 10 \\
        \bottomrule
    \end{tabular}
\end{table}
The desired behavior for the MBR is to rapidly reach and maintain the inverted pose, and regulate the $\psi$ to zero, while minimizing energy consumption.
The hyperparameters used in Algorithms~\ref{al:rl_experience_collection} and~\ref{al:rl_policy_training} are listed in Table~\ref{tab:training_params}.

The desired MBR behavior is to rapidly reach and maintain an inverted pose, followed by adjusting the $\psi$ to zero. At the same time, energy consumption must be minimized, as the reward function includes a penalty for action.

The parameters and their corresponding values used in Algorithms~\ref{al:rl_experience_collection} and~\ref{al:rl_policy_training} are provided in Table \ref{tab:training_params}.
\begin{table}[t]
    \centering
    \setlength{\tabcolsep}{3pt}  
    \caption{Parameters in policy training}
    \label{tab:training_params}
    \begin{tabular}{*{13}c}
        \toprule
        $N$ & $t_e$ & $N_e$ & 
        $\sigma$ & $\xi$ & $n_{\sigma}$ & 
        $d_p$ & $\gamma$ & $\varpi$ & $c_a$ & $c_b$ & $\lambda$ & $\psi_0$\\
        \midrule
        10 & 30s & 500 & 
        0.15 & 0.95 & 100 & 
        2 & 0.98 & 0.01 & 0.1 & 0.1 & [0.6, 1] & 0.5\\
        \bottomrule
    \end{tabular}
\end{table}
Ten buffers were used to store experiences sampled across different values of $\lambda$, which were constrained to the interval $[0.6, 1.0]$, given the MBR parameters outlined in \cite{tao2020design}. 

\section{Evaluation and Experiment}
To evaluate the performance of the learned policy, the parameters $m_w$, $\lambda$, and $g_m$ were varied in the inverted pose stabilization task. The energy-shaping controller proposed in~\cite{wang2024achieving} was adopted as the baseline for comparison. Its control gains were tuned under nominal conditions of $m_w = 23.35$ g, $\lambda = 1$, and $g_m = 1.7$.
The objective was to drive the $\phi$ from $0$ to $\pi$, while maintaining both $\theta$ and $\psi$ at zero.
\begin{figure}[t]
    \centering
    \includegraphics[width=1\linewidth]{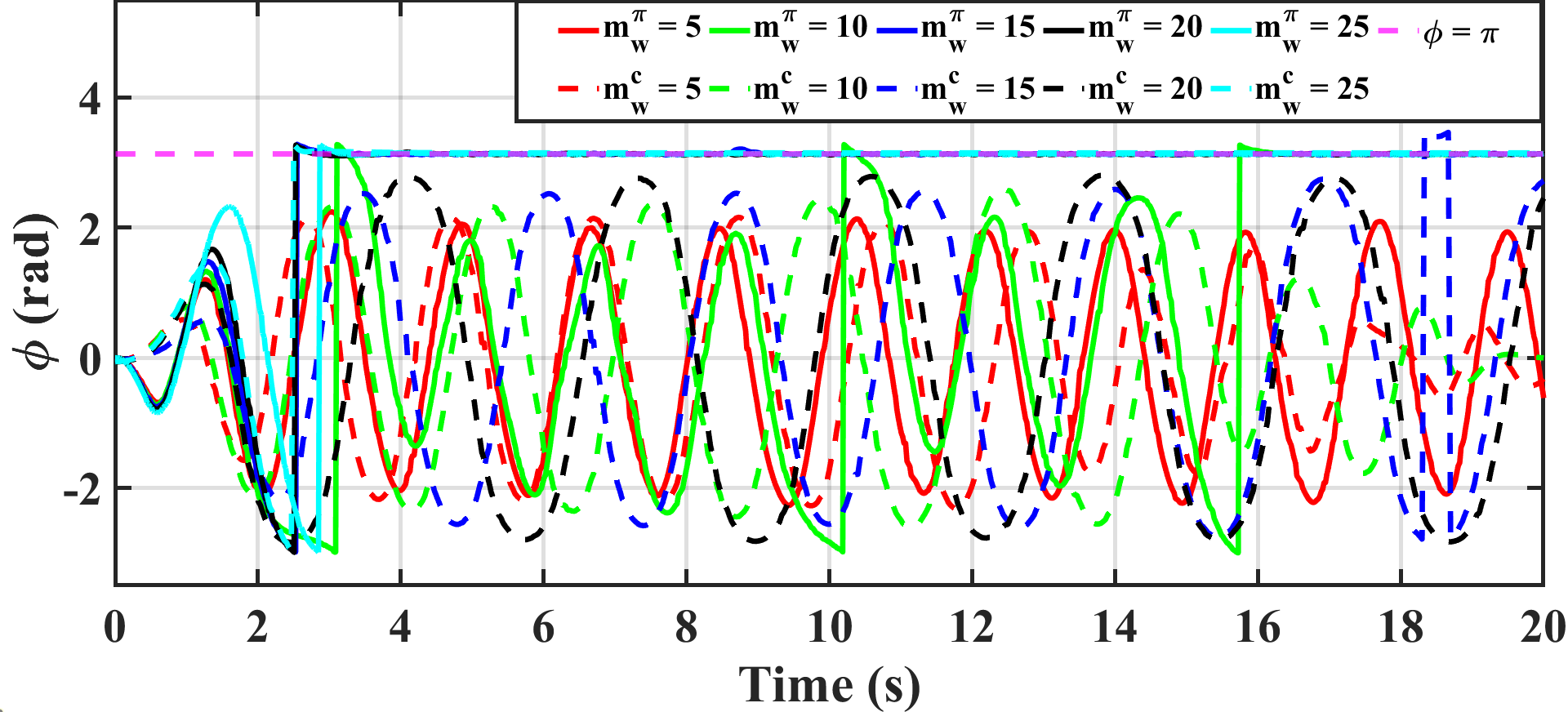}
    \caption{Roll angle variations with changes in $m_w$. Solid lines represent the learned policy, while the dotted line represents the baseline controller.}
    \label{fig:res_m_w}
\end{figure}
\subsection{Performance of the Learned Policy to $m_w$ Variations} \label{st:m_w}
Table~\ref{tab:change_m_w} summarizes the results obtained by varying the $m_w$, with $\lambda = 1$ and $g_m = 1.7$ held constant.
\begin{table}[ht]
    \centering
    \setlength{\tabcolsep}{2pt}  
    \caption{Comparison of the baseline controller and the learned policy with varied $m_w$}
    \label{tab:change_m_w}
    \begin{tabular}{*{1}{c|}*{5}c}
        \toprule
        \textbf{Method} & $m_w = 5g$ & $m_w = 10g$ & $m_w = 15g$ & $m_w = 20g$ & $m_w = 25g$\\
        \midrule
        \textbf{Baseline} &$\times$ & $\times$ & $\times$ & $\times$ & $\checkmark$\\
        \textbf{Our policy} & $\times$ & $\checkmark$ & $\checkmark$ & $\checkmark$ & $\checkmark$\\
        \bottomrule
    \end{tabular}
\end{table}

The parameter $m_w$ was varied from $5$ g to $25$ g, covering conditions from buoyancy-dominant (buoyancy $>$ gravity) to gravity-dominant (buoyancy $<$ gravity). When $m_w = 5$ g, neither the learned policy nor the baseline controller completed the task. In this case, $c_g$ is too close to $c_t$, preventing the generation of sufficient rotational moment to invert the MBR under the fixed motor gain $g_m = 1.7$.
For all other values of $m_w$, the learned policy successfully achieved the inverted pose, whereas the baseline controller was effective only at $m_w = 25$ g$, $ its nominal tuning condition. These results suggest that the baseline controllers are sensitive to parameter variations, while the learned policy exhibits stronger robustness across different dynamic configurations.
The $\phi$ responses are shown in Fig.~\ref{fig:res_m_w}, where $m_w^{\pi}$ and $m_w^{c}$ denote the policy and baseline controller, respectively. 

As $m_w$ increases, the maximum achievable $\phi$ increases and eventually reaches $\pi$. The neutrally buoyant weight is approximately 23.35 g; thus, at $m_w = 25$ g, gravity exceeds buoyancy. In this regime, $c_g$ moves closer to $c_b$, reducing the restoring moment and allowing a larger achievable $\phi$.

\subsection{Performance of the Learned Policy to $\lambda$ Variations} \label{st:res_lambda}
Varying $\lambda$ determines the position of $c_g$ while maintaining the MBR in a neutrally buoyant state. In this experiment, the extra weight and motor gain were set to $m_w = 23.35$ g and $g_m = 1.7$, respectively. Table~\ref{tab:change_lambda} presents the results for both the baseline controller and the learned policy, indicating that the baseline controller was only successful when $\lambda = 1.0$.
\begin{table}[t]
    \centering
    \setlength{\tabcolsep}{4pt}  
    \caption{Comparison of the baseline controller and the learned policy with varied $\lambda$}
    \label{tab:change_lambda}
    \begin{tabular}{*{1}{c|}*{5}c}
        \toprule
        \textbf{Method} & $\lambda = 0.6$ & $\lambda = 0.7$ & $\lambda = 0.8$ & $\lambda = 0.9$ & $\lambda = 1.0$\\
        \midrule
        \textbf{Baseline} & $\times$ & $\times$ & $\times$ & $\times$ & $\checkmark$\\
        \textbf{Our policy} & $\checkmark$ & $\checkmark$ & $\checkmark$ & $\checkmark$ & $\checkmark$\\
        \bottomrule
    \end{tabular}
\end{table}

\begin{figure}[t]
    \centering
    \includegraphics[width=1\linewidth]{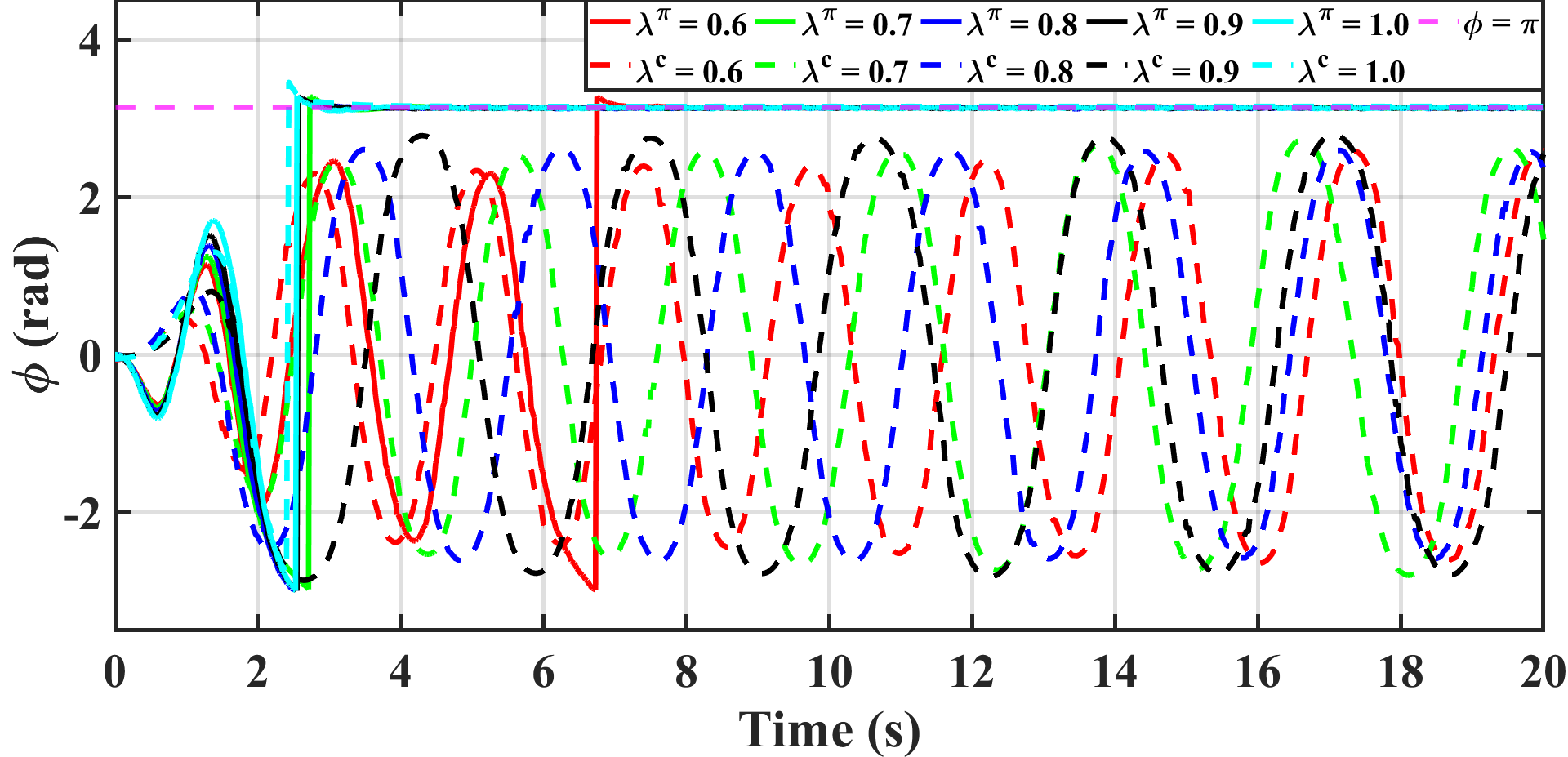}
    \caption{Roll angle variations with changes in $\lambda$.}
    \label{fig:res_lambda}
\end{figure}

The trained control policy completes the task for all $\lambda \in [0.6, 1.0]$, demonstrating its robustness against variations in MBR parameters. The variations in the roll angle under these conditions are depicted in Fig.~\ref{fig:res_lambda}, in which the parameter for the policy is denoted as $\lambda^{\pi}$ and that for the controller as $\lambda^c$.
Unlike the variation in $m_w$, changes in $\lambda$ do not affect the magnitudes of buoyancy and gravity, but only influence the position of $c_g$. As $\lambda$ increases, $c_g$ moves closer to $c_b$. This results in larger maximum achievable roll angles (before reaching $\phi = \pi$) and reduces the time required to achieve the inverted state. When $\lambda = 1.0$, the controller completed the task more quickly, as its parameters were specifically fine-tuned for this configuration.

\subsection{Performance of the Learned Policy to $g_m$ Variations}
To verify that the learned policy functions effectively across different motors, $m_w$ and $\lambda$ are configured to 23.35 g and 1.0, respectively. The results are shown in Table~\ref{tab:change_gm}.
\begin{table}[t]
    \centering
    \setlength{\tabcolsep}{3.5pt}  
    \caption{Comparison of the baseline controller and the learned policy with varied $g_m$}
    \label{tab:change_gm}
    \begin{tabular}{*{1}{c|}*{5}c}
        \toprule
        \textbf{Method} & $g_m = 0.5$ & $g_m = 1.0$ & $g_m = 1.5$ & $g_m = 2.0$ & $g_m = 2.5$\\
        \midrule
        \textbf{Baseline} & $\times$ & $\checkmark$ & $\checkmark$ & $\checkmark$ & $\checkmark$\\
        \textbf{Our policy} & $\times$ & $\checkmark$ & $\checkmark$ & $\checkmark$ & $\checkmark$\\
        \bottomrule
    \end{tabular}
\end{table}

\begin{figure}[t]
    \centering
    \includegraphics[width=1\linewidth]{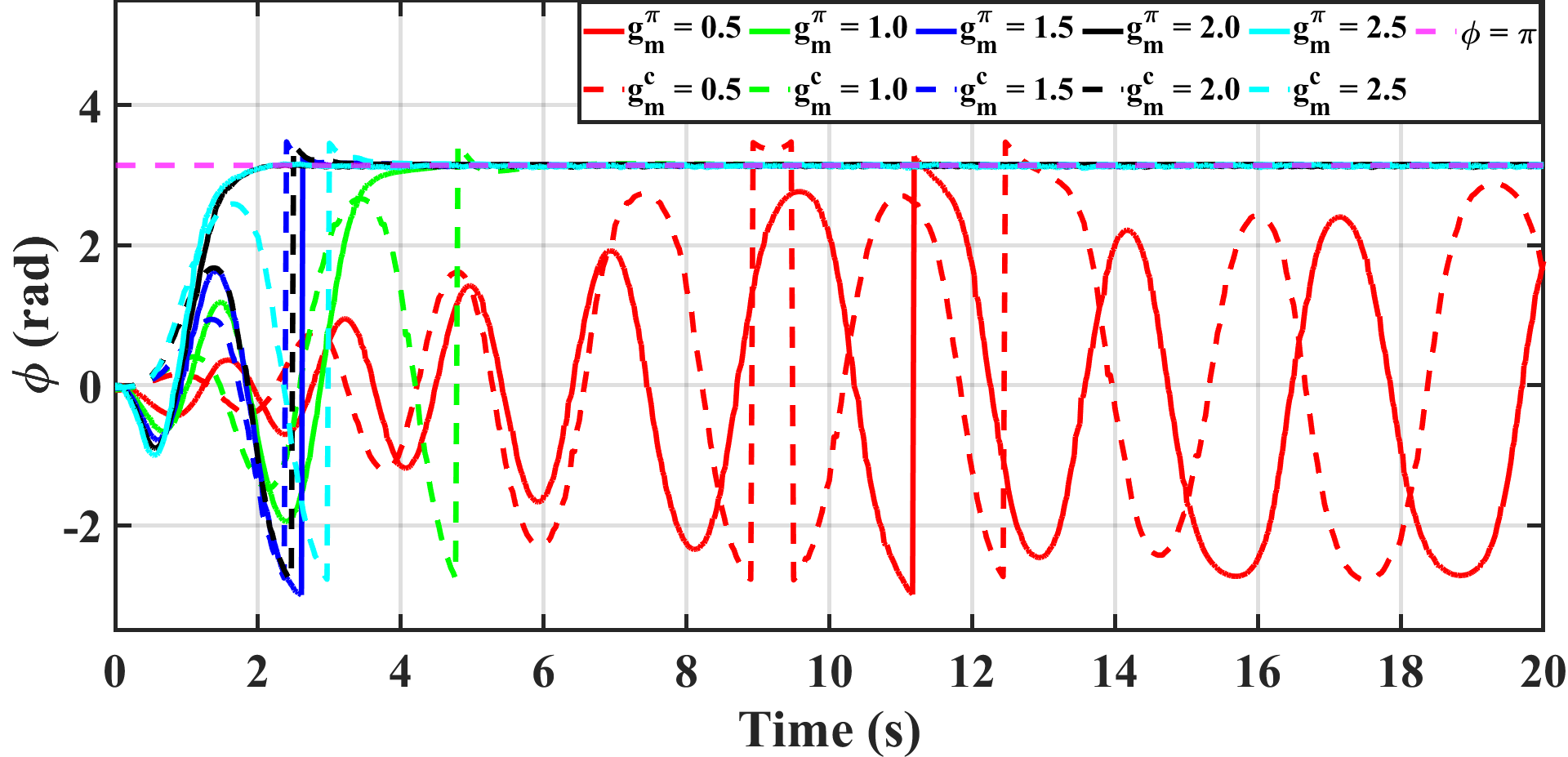}
    \caption{Roll angle variations with changes in $g_m$.}
    \label{fig:res_gm}
\end{figure}

Only when $g_m = 0.5$ did both the learned policy and the baseline controller fail. Although each method could temporarily drive the MBR to the inverted pose, the motor thrust was insufficient to maintain it. The controller achieved inversion at approximately 9 s, while the policy did so at around 11 s, but both subsequently lost stability.
For all other tested values of $g_m$, both the baseline controller and the learned policy successfully completed the task. However, their control behaviors differ noticeably, as illustrated in Fig.~\ref{fig:res_gm}.
The results indicate that as $g_m$ increases, the policy requires less time to complete the task, owing to the corresponding increase in total force. When $g_m \ge 2.0$, only two rotations are necessary, a result consistent with the controller method. In contrast, for the controller, a higher $g_m$ results in a larger maximum achievable roll angle, which in turn requires more time to complete the task.
\begin{table}[h]
    \centering
    \setlength{\tabcolsep}{12.25pt}
    \caption{Comparison of the baseline controller and the learned policy with varied $m_w$, $\lambda$, and $g_m$}
    \label{tab:change_m_w_lambda_g_m}
    \begin{tabular}{*{1}{c|}*{5}c}
        \toprule
        \textbf{Test Case} & \textbf{C1} & \textbf{C2} & \textbf{C3} & \textbf{C4} & \textbf{C5} \\
        \midrule
        \quad $m_w$ (\si{\gram}) & 15 & 15 & 20 & 25 & 25 \\
        \quad $\lambda$ & 0.8 & 1.0 & 0.9 & 0.7 & 0.8 \\
        \quad $g_m$ & 1.7 & 1.0 & 1.6 & 1.5 & 1.4 \\
        \midrule
        \textbf{Baseline} & $\times$ & $\times$ & $\times$ & $\times$ & $\times$ \\
        \textbf{Our Policy} & $\checkmark$ & $\checkmark$ & $\checkmark$ & $\checkmark$ & $\checkmark$ \\
        \bottomrule
    \end{tabular}
\end{table}

\begin{figure}[h]
    \centering
    \includegraphics[width=1\linewidth]{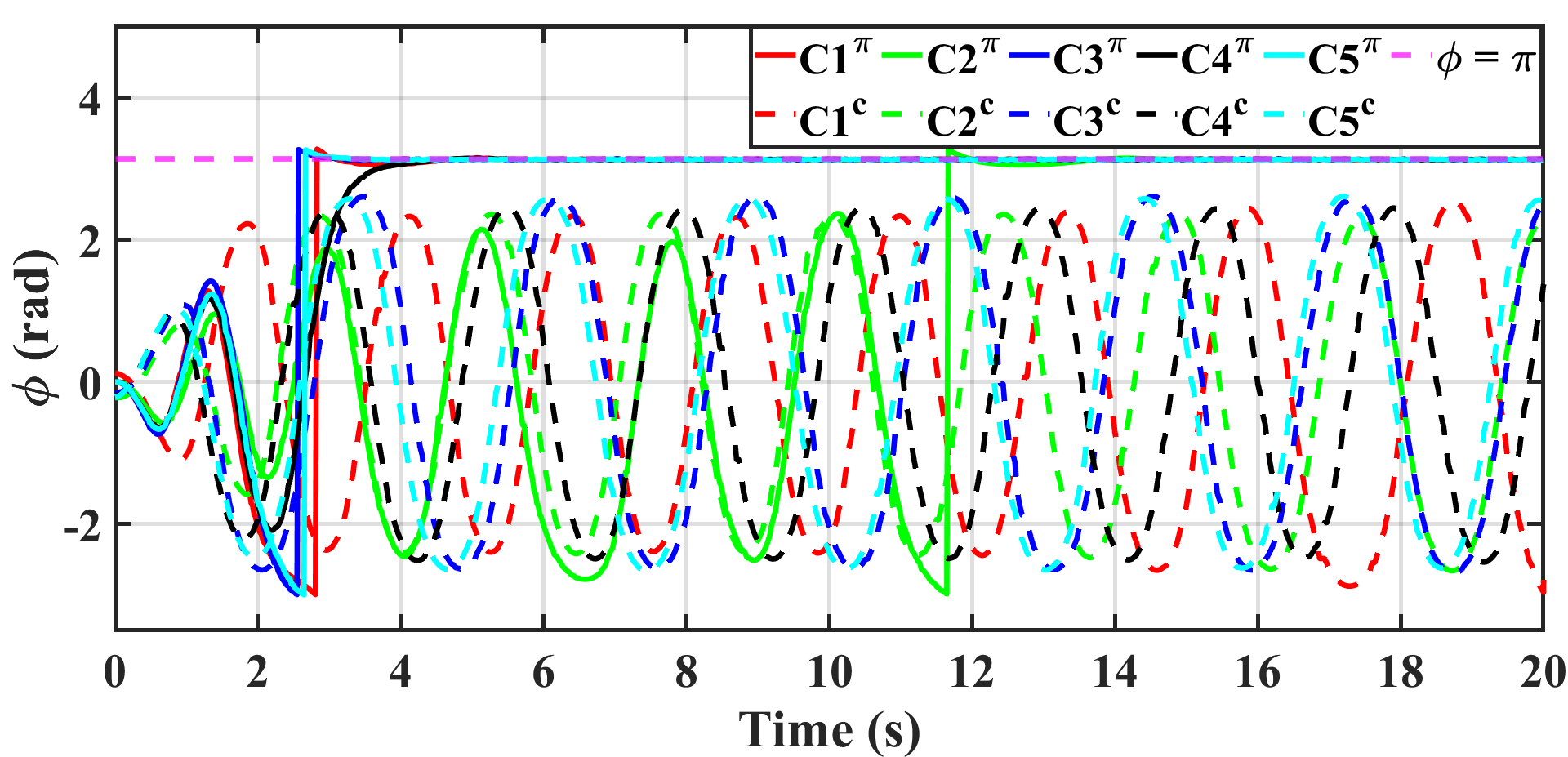}
    \caption{Roll angle variations with changes in $m_w$, $\lambda$ and $g_m$.}
    \label{fig:res_gm_lambda_m_w}
\end{figure}
\subsection{Performance of the Learned Policy to $m_w$, $\lambda$ and $g_m$ Variations}
The parameters $m_w$, $\lambda$, and $g_m$ were varied simultaneously to further evaluate the robustness of the policy. The configurations and results are summarized in Table~\ref{tab:change_m_w_lambda_g_m}. The learned policy succeeded in all cases, whereas the baseline controller failed under these combined variations.
The $\phi$ responses are shown in Fig.~\ref{fig:res_gm_lambda_m_w}. In Case 2 (C2), the maneuver required a longer completion time due to the low motor gain $g_m$, which limited the available actuation capability.

\subsection{Ablation Study}
\begin{figure}[t]
    \centering
    \includegraphics[width=1\linewidth]{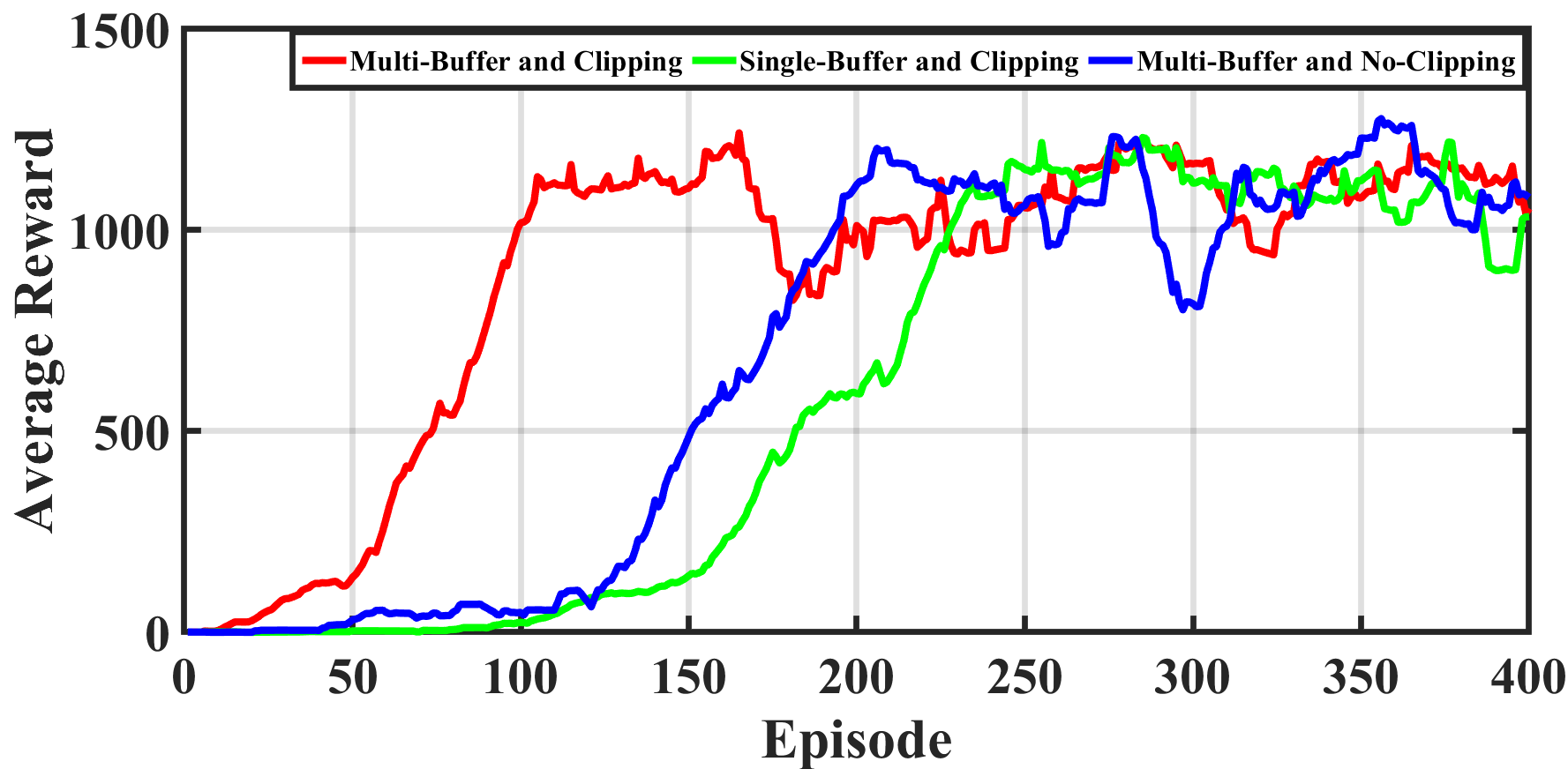}
    \caption{Comparison of the average reward across three cases, computed using a moving average with a window size of 19.}
    \label{fig:res_reward}
\end{figure}

To evaluate the contribution of the multi-buffer strategy and gradient clipping to training stability, an ablation study was conducted. The results are shown in Fig.~\ref{fig:res_reward}.
The proposed method, which combines multi-buffer experience storage with gradient clipping, converged within approximately 100 episodes. Removing gradient clipping while retaining the multi-buffer structure increased the convergence time to nearly 200 episodes. In contrast, using a single buffer with gradient clipping required at least 250 episodes—2.5 times slower than the proposed approach. For fairness, the capacity of the single buffer was set equal to the total capacity of all buffers in the multi-buffer configuration.
These results demonstrate that the combination of multi-buffer sampling and gradient clipping significantly improves training stability and sample efficiency. The average return exhibits persistent fluctuations due to the continuous injection of exploration noise $\epsilon_t^k$ throughout training.

\subsection{Policy Deployment in a Physical MBR}
\begin{figure}[t]
    \centering
    \includegraphics[width=1\linewidth]{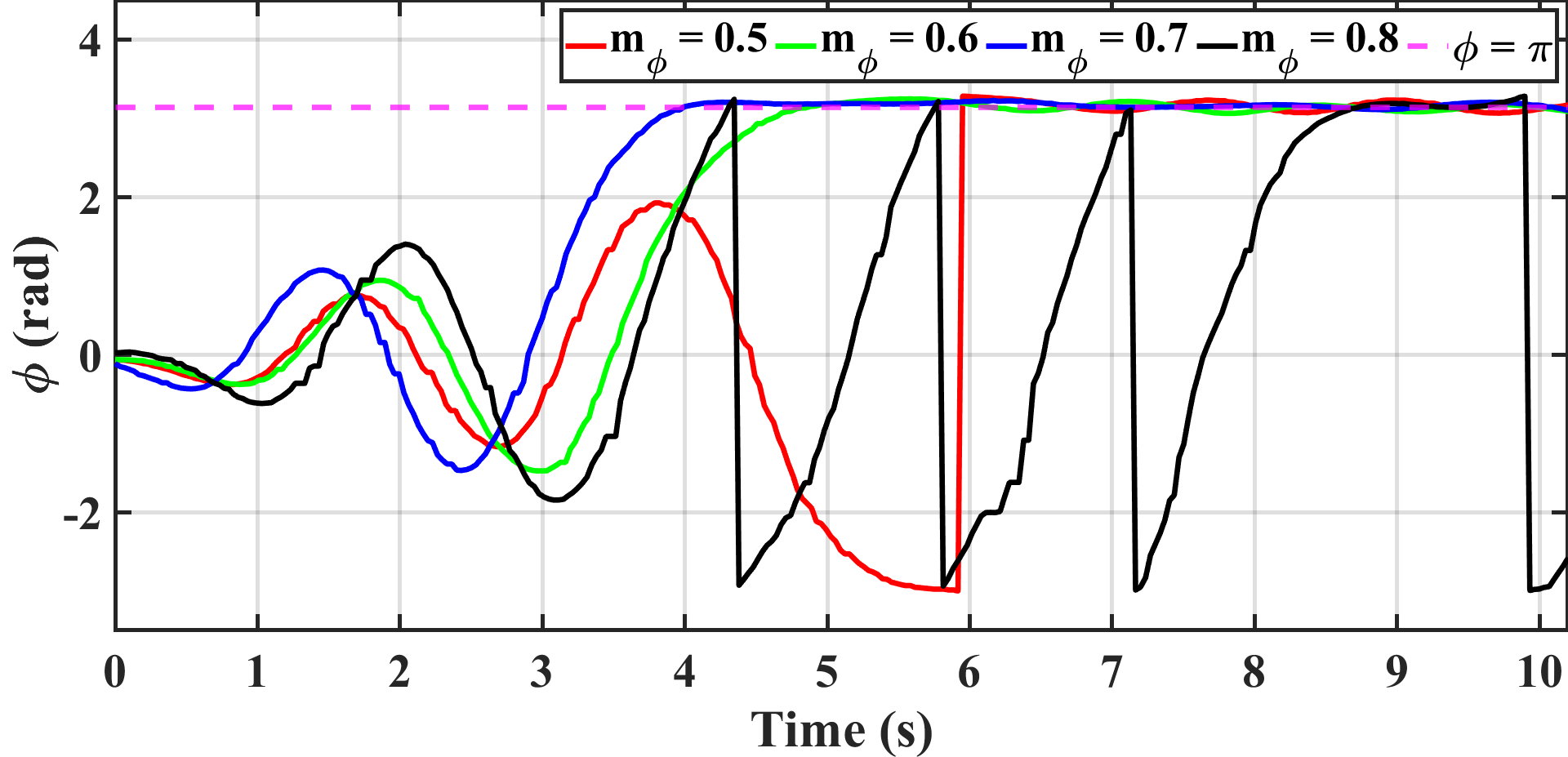}
    \caption{Roll angle variations of the MBR with different $m_\phi$.}
    \label{fig:res_mapping_layer}
\end{figure}

\begin{figure}[ht]
    \centering
    \includegraphics[width=1\linewidth]{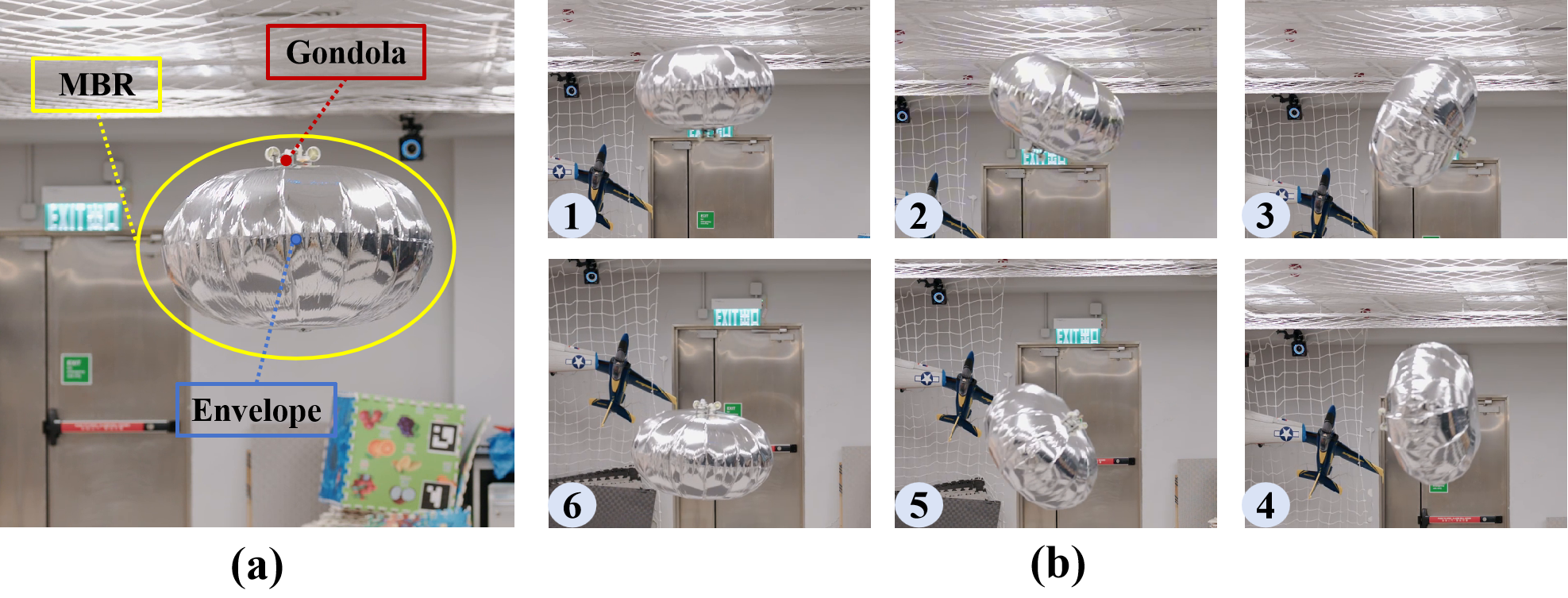}
    \caption{(a) Experimental setup; (b) Visualization of the action sequence of an MBR for achieving an inverted pose.}
    \label{fig:phy_res_0.7}
\end{figure}

\begin{figure}[h]
    \centering
    \includegraphics[width=1\linewidth]{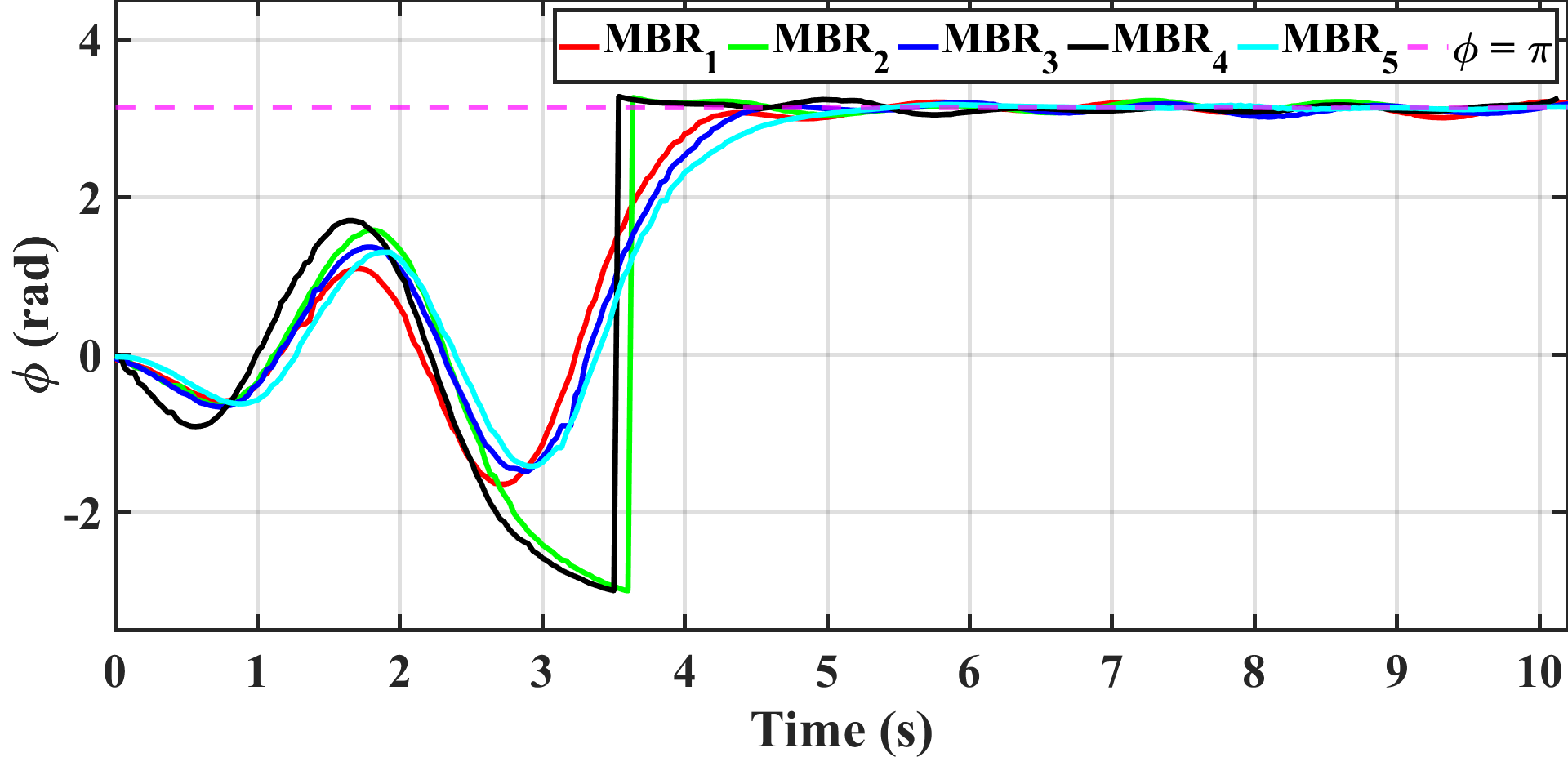}
    \caption{Roll angle variation of the MBR with different $m_w$.}
    \label{fig:phy_res_weights}
\end{figure}

We transfer the learned policy to the real platform with only minimal parameter adjustments, avoiding additional training on physical data.
As shown in Fig.~\ref{fig:system_framework}, a mapping layer is introduced to mitigate the sim-to-real discrepancy during the inverted transition:
\begin{equation}
\mathbf{\tau}^{p} = \mathbf{M}_0 \boldsymbol{a}^p,
\quad \Delta{\phi} < \varrho,
\end{equation}
where $\mathbf{\tau}^{p} = [\tau_x, \tau_y, \tau_z]^T$ denotes the physical torque command. The term $\Delta_{\phi} = \pi - \phi$ represents the roll angle deviation, and $\varrho$ is the switching threshold. In experiments, $\mathbf{M}_0 = \mathrm{diag}(m\phi, m_\theta, m_\psi)$ with $\varrho = 0.8$. The parameters $m_\theta$ and $m_\psi$ are fixed at 0.1, while $m_\phi$ varies from 0.5 to 0.8.
The results are shown in Fig.~\ref{fig:res_mapping_layer}. The learned policy drives the MBR to the inverted pose, after which a PD controller stabilizes the system once angular velocities approach zero. The transition sequence for $m_\phi = 0.7$ is illustrated in Fig.~\ref{fig:phy_res_0.7}. Among the tested values, only $m_\phi = 0.8$ failed. These results indicate that the proposed mapping layer effectively bridges the sim-to-real gap without policy retraining.
Using $m_\phi = 0.7$, additional physical experiments were conducted by varying $m_{w_1}$ and $m_{w_2}$ (Table~\ref{tab:change_m1_m2}). The MBR successfully achieved inversion in all cases (Fig.~\ref{fig:phy_res_weights}). Increasing $m_{w_1}$ shifts the $c_g$ toward the $c_b$, reducing transition time, whereas increasing $m_{w_2}$ moves $c_g$ toward the $c_t$, prolonging the maneuver.
These observations are consistent with the simulation results in Sections~\ref{st:m_w} and~\ref{st:res_lambda}, further validating the effectiveness of the proposed method.
\begin{table}[h]
    \centering
    \setlength{\tabcolsep}{7pt}  
    \caption{Configuration of the extra weights in the physical experiment}
    \label{tab:change_m1_m2}
    \begin{tabular}{*{1}{c|}*{5}c}
        \toprule
        \textbf{Weights} & $MBR_1$ & $MBR_2$ & $MBR_3$ & $MBR_4$ & $MBR_5$\\
        \midrule
        \textbf{$m_{w1}(g)$} & $25$ & $26.07$ & $25$ & $27.59$ & $25$ \\
        \textbf{$m_{w2}(g)$} & $0$ & $0$ & $1.07$ & $0$ & $2.59$\\
        \bottomrule
    \end{tabular}
\end{table}

\section{Conclusion}
This paper proposes a new DRL-based method for inverted control of MBR, aiming to achieve its full agility. The method involves the construction of a virtual training environment, policy training using domain randomization, an improved TD3 method, and policy deployment via a designed mapping layer. Compared to the energy-shaping controller, the learned policy achieves a higher success rate across diverse scenarios. Although the mapping layer designed for policy deployment enables the policy to function in physical settings without further training, it constrains the performance of the learned policy. This indicates that a linear relationship alone cannot fully bridge the sim-to-real gap. Therefore, analyzing and quantifying the sim-to-real gap in inverted control remains an open problem for future work.

\bibliographystyle{IEEEtran}
\bibliography{ref}

\end{document}